\documentclass[pdflatex,sn-mathphys-num]{sn-jnl}% Math and Physical Sciences Numbered Reference Style
\usepackage{graphicx}%
\usepackage{multirow}%
\usepackage{amsmath,amssymb,amsfonts}%
\usepackage{amsthm}%
\usepackage{mathrsfs}%
\usepackage[title]{appendix}%
\usepackage{xcolor}%
\usepackage{textcomp}%
\usepackage{manyfoot}%
\usepackage{booktabs}%
\usepackage{algorithm}%
\usepackage{algorithmicx}%
\usepackage{algpseudocode}%
\usepackage{listings}%
\theoremstyle{thmstyleone}%
\theoremstyle{thmstyletwo}%

\theoremstyle{thmstylethree}%

\begin{document}

\title[Article Title]{$ $ $ $ $ $ $ $ $ $ Improving AI Efficiency in Data Centres \\by Power Dynamic Response}


\author*[1]{\fnm{Andrea} \sur{Marinoni}}\email{am2920@cam.ac.uk}

\author[2]{\fnm{Sai} \sur{Shivareddy}}\email{sai.shivareddy@nyobolt.com}

\author[1]{\fnm{Pietro} \sur{Lio'}}\email{pl219@cam.ac.uk}

\author[3]{\fnm{Weisi} \sur{Lin}}\email{wslin@ntu.edu.sg}

\author[3]{\fnm{Erik} \sur{Cambria}}\email{cambria@ntu.edu.sg}

\author[4]{\fnm{Clare P.} \sur{Grey}}\email{cpg27@cam.ac.uk}


\affil*[1]{\orgname{Department of Computer Science and Technology, University of Cambridge}, \orgaddress{\street{15 JJ Thomson ave.}, \city{Cambridge}, \postcode{CB3 0FD}, \country{United Kingdom}}}

\affil[2]{\orgname{Nyobolt Limited}, \orgaddress{\street{Evolution Business Park, Unit 2}, \city{Cambridge}, \postcode{CB24 9NG}, \country{United Kingdom}}}

\affil[3]{\orgname{College of Computing and Data Science, Nanyang Technological University}, \orgaddress{\street{50 Nanyang Ave}, \postcode{639798}, \country{Singapore}}}

\affil[4]{\orgname{Yusuf Hamied Department of Chemistry}, \orgaddress{\street{Lensfield Rd}, \city{Cambridge}, \postcode{CB2 1EW}, \country{United Kingdom}}}

%\affil[2]{\orgdiv{Department}, \orgname{Organization}, \orgaddress{\street{Street}, \city{City}, \postcode{10587}, \state{State}, \country{Country}}}

%\affil[3]{\orgdiv{Department}, \orgname{Organization}, \orgaddress{\street{Street}, \city{City}, \postcode{610101}, \state{State}, \country{Country}}}


\abstract{The steady growth of artificial intelligence (AI) has accelerated in the recent years, facilitated by the development of sophisticated models such as large language models and foundation models. 
Ensuring robust and reliable power infrastructures is fundamental to take advantage of the full potential of AI.  
However, AI data centres are extremely hungry for power, putting the problem of their power management in the spotlight, especially with respect to their impact on environment and sustainable development.  
In this work, we investigate the capacity and limits of solutions based on an innovative approach for the power management of AI data centres, i.e., making part of the input power as dynamic as the power used for data-computing functions. 
%This strategy identifies a paradigm shift in the AI data centre power management, that could 
The performance of passive and active devices are quantified and compared in terms of computational gain, energy efficiency, reduction of capital expenditure, and management costs by analysing power trends from multiple
data platforms worldwide. 
This strategy, which identifies a paradigm shift in the AI data centre power management, has the potential to strongly improve the sustainability of AI hyperscalers, enhancing their footprint on environmental, financial, and societal fields.

%reducing the stress on power management imposed
%by AI models workload becomes essential to enable the data centre growth and ensure
%the effectiveness of their investments.

%The development and uptake of artificial intelligence (AI) has accelerated in recent years – elevating the question of what widespread deployment of the technology will mean for the energy sector. There is no AI without energy – specifically electricity for data centres. At the same time, AI could transform how the energy industry operates if it is adopted at scale. However, until now, policy makers and other stakeholders have often lacked the tools to analyse both sides of this issue due to a lack of comprehensive data. 

%This report from the International Energy Agency (IEA) aims to fill this gap based on new global and regional modelling and datasets, as well as extensive consultation with governments and regulators, the tech sector, the energy industry and international experts. It includes projections for how much electricity AI could consume over the next decade, as well as which energy sources are set to help meet it. It also analyses what the uptake of AI could mean for energy security, emissions, innovation and affordability.
}




\maketitle

\section{AI hunger for power}\label{sec1}

The solid advancements of microelectronics technology that occurred in the last decades enabled the emergency of high performance data analysis systems in every field of science~\cite{Alberta,PowerAI}. 
The gains in computing capabilities as well as the improvement of chip efficiency (both in terms of scalability and storage) have facilitated the development of advanced artificial intelligence systems like large-language models (LLMs) and other foundation models. 
These schemes allowed us to extract information across diverse operational scenarios (from environmental monitoring to proteomics, from financial market analysis to robotics and automotive industry) by processing extreme volumes of records acquired by multiple sources of information~\cite{Alberta,PowerAI,NPJ_water,NatClimChange_Masanet,NatClimChange_Rolnick,Science_masanet}. 
To achieve accurate and robust information extraction, high performance AI requires to access in the scale of milliseconds to terabytes of data in data centres and AI hyperscalers located across the world~\cite{Verne_AIheat,JLL,McKinsey}. 
This is expected to generate a huge economic value throughout the global economy, that is estimated to fall between 2.6 and 4.4 trillion USD annually.~\cite{Alberta,McKinsey,JLL,McKinsey2}

%\color{blue}
%\textbf{cite MCKINSEY ``SCALING BIGGER FASTER CHEAPER DATA CENTRES"}
%\color{black}

It is not surprising then that power management plays a key role in AI investments and development. 
Indeed, power infrastructure is crucial to ensure that the potential of AI can be fully realised~\cite{JLL,AIEnergyConsum,DataCentre_footprint,DataCentre_footprint2,DataCentreDilemma,PowerHungry,Science_masanet}. 
In fact, AI data centres are hungry for power, with figures that have a direct impact on sustainability of energy production as well as efficiency. 
For instance, it has been estimated 
that each additional data centre in the next five years would require between 50 and 60GW, leading to an investment of more than 500 billion USD in data centre infrastructure in the United States alone~\cite{McKinsey}. 
This translates into an increase by approximately 400 TWh 
between 2024 and 2030 for the electricity demand for data centres, which reflects a compound annual growth rate (CAGR) of circa 23$\%$~\cite{McKinsey,McKinsey2}.

It is important to consider that power management of AI data centres can face a number of structural limitations~\cite{PowerAI,Verne_AIheat,Alberta,McKinsey,MSOpenAINvidia,OpenAI_compute,SustainableAI,SuperComputing}. 
In particular, identifying reliable and sustainable power sources, as well as guaranteeing upstream infrastructures for power access, becomes a cumbersome exercise. %, especially considering the need for data centre growth. 
This is especially true in areas where grid access can be made complicated by lack of power equipment, reduced electrification, and aging power plants. 
As such, reducing the stress on power management imposed by AI models workload becomes essential to enable the data centre growth and ensure the effectiveness of their investments. 
Specifically, AI workloads are characterized by~\cite{Alberta,TAPAS_Microsoft,Microsoft24,MSOpenAINvidia,DataCentre_footprint,DataCentreDilemma,AIEnergyConsum,OpenAI_compute,NatClimChange_Rolnick,NatClimChange_Masanet,SustainableAI,GPUEnergyConsum,Brookhaven,LLM_megatron}:
\begin{itemize}
    \item high computational intensity over long timeframes;
    \item high degree of variability, unpredictability, and nonlinear scalability of computational power usage;
    \item sensitivity to algorithmic design and implementation.
\end{itemize}

%\begin{figure}[h]
%\centering
%\includegraphics[width=0.5\textwidth]{FC5412 - Nyobolt - Infographics White Paper-V2[64]_Page_3_2.jpg}
%\caption{Potential issues for demand, supply and impact caused by }
%\label{fig_impact}
%\end{figure}

Failing to consider these aspects when designing, developing and implementing AI hyperscalers and data centres leads to catastrophic disruption of the service, which can be grouped (according to OECD categories) with respect to the demand and supply of AI data centres, as well as their impact on society. Specifically~\cite{McKinsey,JLL,Verne_AIheat,TAPAS_Microsoft,Alberta,McKinsey2, IEA,Bloomberg,NatClimChange_Masanet,NatClimChange_Rolnick,DataCentre_footprint,DataCentreDilemma,LCA_AI,NPJ_water,PowerHungry,Science_masanet,GPUEnergyConsum,SustainableAI,SuperComputing}:
\begin{itemize}
    \item \textbf{demand}: the unpredictable number of users accessing the data centre platforms as well as the variable load of the various jobs run over the data centre architecture translates in high randomness of the usage of AI accelerators (e.g., graphics processing units (GPUs), tensor processing units (TPUs)). When the stress on the power grid exceeds the structural limits of the given data centres, the access to AI accelerators can be discontinued, hence resulting in interruptions of the AI analysis service;
    \item \textbf{supply}: the aforesaid AI accelerator shut down would mean the data centres failing to comply to their functions, hence making the structural investments ineffective or void. To avoid this problem, AI data centres managers tend to oversize the grid connections, power distribution units (PDUs), and backup systems.  
    This hence imposes additional financial effort to support the data centre demand, and keep energy demand constant;
    \item \textbf{impact}: the vast data centre infrastructures required to ensure robust and reliable data analysis result in a high impact on environment and sustainability. It has been estimated that each MW of server power produces 1.3 MW of heat released in the atmosphere. Also, the high power comsumption and variable load of data processing make data centres affect the grid stability of entire regions, this affecting key welfare and socioeconomic factors of local, regional and national communities and governments. 
\end{itemize}

\section{A paradigm shift} 

\begin{figure}[h]
\centering
\includegraphics[width=1\textwidth]{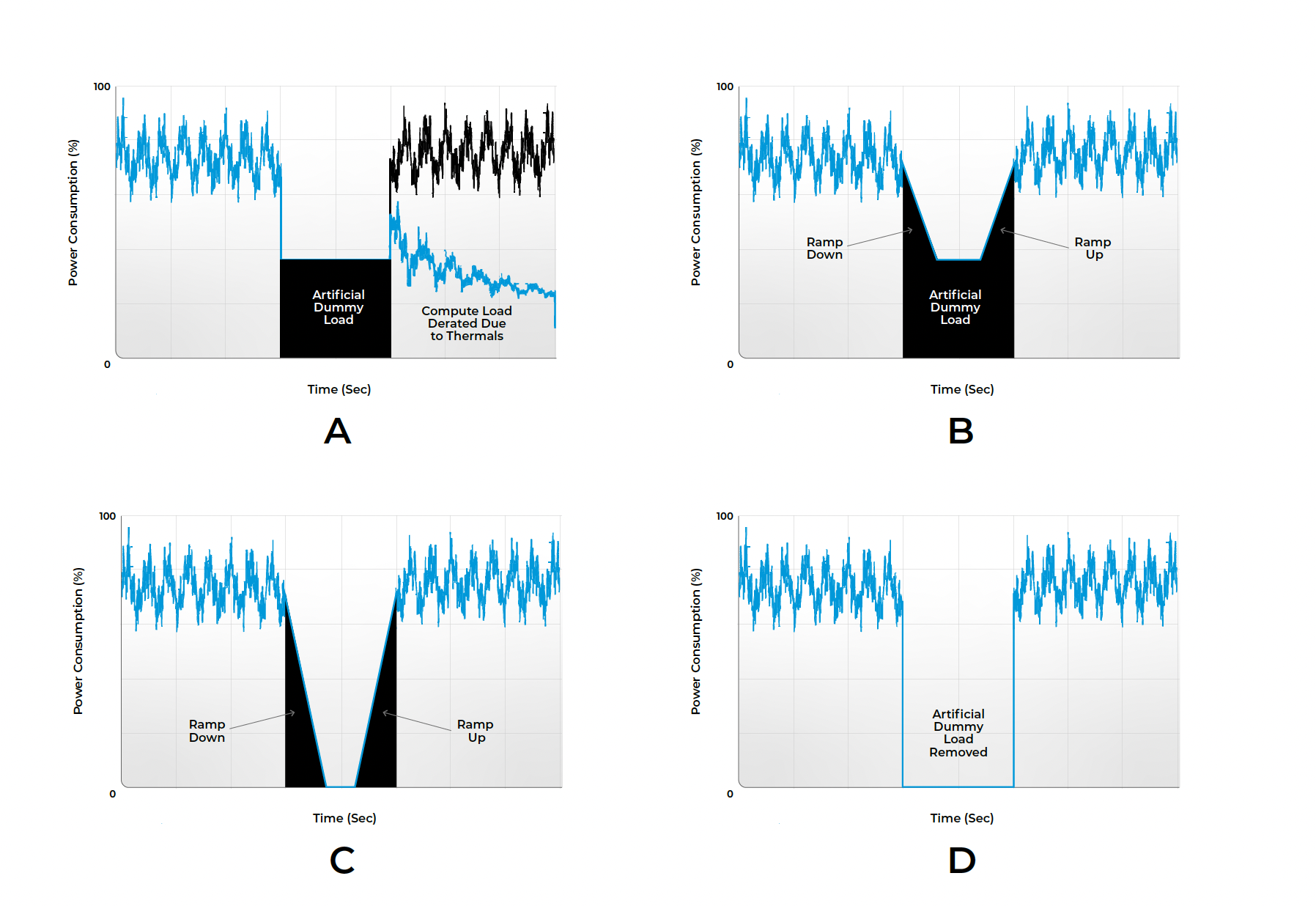}
\caption{Typical trends of AI accelerator power draw (light blue line) through time. (A): State-of-the-art approach: the dummy loads (black shaded area) are used during idle intervals to reduce the amplitude of the power fluctuations. The use of dummy loads leads to a degradation of computational load (with respect to the required power profiles - in black line), because dummy loads deteriorate the thermal profiles of AI accelerators. (B-D): Power trends when solutions for dynamic power response are employed. (B,C): Power trends profiles when passive devices are used. (D): Power trends profiles when actives devices are used.}
\label{fig_Dummy}
\end{figure}

The aforementioned problems result from the inability of input power structures to track and follow the high variability of power profiles induced by AI models use. 
The current solution for this relies on the implementation of artificial ``dummy loads'' that run between actual AI accelerators compute cycles (Figure~\ref{fig_Dummy}(A)). 
These artificial compute loads are used between real compute cycles, and are typically used to avoid sharp fluctuations in grid draw~\cite{Nvidia_capacitor,Skeleton_GPUGraphene,Alberta,Microsoft24,TAPAS_Microsoft}. 
On one hand, this approach leads to 
using more energy, generating excess heat which introduces thermal de-rating of AI accelerators that reduces their compute capacity~\cite{Alberta,TAPAS_Microsoft,Skeleton_GPUGraphene}. 
On the other hand, this solution leads to inflated capital expenditure (CAPEX), underutilized infrastructure (to be estimated in the order of billions of dollars globally, and millions of dollars per data centre), and additional grid connection delays~\cite{McKinsey,JLL}.

Analytically, the power balance of AI data centres can be written as follows~\cite{Alberta}:

\begin{equation}
    \underbrace{P_{grid} + P_{ext}}_{Input} = \underbrace{P_{infra} + P_{comp}}_{Output},
    \label{eq_power_balance}
\end{equation}

\noindent where the four terms identify the following:
\begin{itemize}
    \item $P_{grid}$: power from distribution grids;
    \item $P_{ext}$: power from external sources;
    \item $P_{infra}$: power used for infrastructures (e.g., cooling, lighting);
    \item $P_{comp}$: power used for data computing-related functions (e.g., AI models, access to memory storage). 
\end{itemize}

In this system, $P_{grid}$, $P_{ext}$, and $P_{infra}$ represent fixed factors, or at least contributions that show very slow dynamics. 
On the other hand, $P_{comp}$ is instead a highly dynamic term, according to the characteristics of AI processing that have been previously introduced. 

This mismatch is at the core of the unique challenges for grid operators when managing  diverse activities in the data centers. 
In fact, physical limitations in the power fluctuations and demand ramp rates leads to service interruptions that could span from one minute to ninety minutes.
%if power demand suddenly ramps up, it can take one minute to 90 minutes for generation resources to respond because of . 
Moreover, the power infrastructures could be put under severe stress by repeating power transients. 
Also, it is worth noting that sudden reduction in power consumption of data centres would result in energy production systems with no outlet for use. 
This affects the sustainability of the data centres and ultimately the regional energy consumption, since  other grid customers can feel the effect of AI data centres power consumptions and fluctuations as spikes or drops in supplied voltage~\cite{McKinsey,JLL,Verne_AIheat,McKinsey2,MSOpenAINvidia}. 
%If the data center suddenly reduces its power consumption, the energy production systems find themselves with excess energy and no outlet.

For these reasons, it is crucial to address the dynamics of the power balance in (\ref{eq_power_balance}) to avoid the occurrence of dramatic events on AI data centres management and infrastructures, as well as to improve their efficiency and effectiveness. 
Indeed, implementing power response units to make the $P_{ext}$ term highly dynamic might enable multiple options for efficient and effective AI power management in data centres~\cite{AIEnergyConsum,OpenAI_compute,MSOpenAINvidia,Alberta,DataCentreDilemma,SustainableAI}: 
\begin{itemize}
    \item AI-aware power pattern modelling through diverse phases (training, fine tuning, inference); 
    \item power ramping/decline compensation; 
    \item protection to overheating; 
    \item adaptive load distribution by means of power/load/temperature scheduling across AI accelerators. 
\end{itemize}

This identifies a paradigm shift that paves the way to a dramatic enhancement of the AI data centre management and effectiveness of their supercomputing performance. 
Also, it enables the design of more robust and successful green AI architectures, as well as improve their environmental impact and carbon footprint~\cite{Alberta,Microsoft24,SustainableAI,DataCentre_footprint,NatClimChange_Rolnick,IEA}.

\section{Results and discussion} 

To compute the potential of this novel approach, it is important to analyze the distribution of power spikes in data centre racks. 
Moreover, the energy contained in these spikes would unveil the measure of the impact of dynamic power response on the current AI data centre management conditions, as well as their perspective growth in the next decades. 
To this aim, we have investigated the AI power trends from multiple data platforms worldwide, focusing our attention on the power fluctuations that each AI accelerator would experience, and considering their effect on the AI data centres as a whole~\cite{Alberta,MIT_supercloud,Microsoft24,TAPAS_Microsoft,Google_dataset,MSOpenAINvidia,Brookhaven}. 

In this respect, it is important to identify the power spikes that occur in these real-life power trends. 
We therefore investigated the aforementioned datasets by moving a threshold across the power draw range: in this approach, every burst of data points that are continuously above this threshold would identify a power spike.
Throughout this work, the said value would define the amount of power that an AI data centre system could absorb by using the power grid source.

\begin{figure}[h]
\centering
\includegraphics[width=1\textwidth]{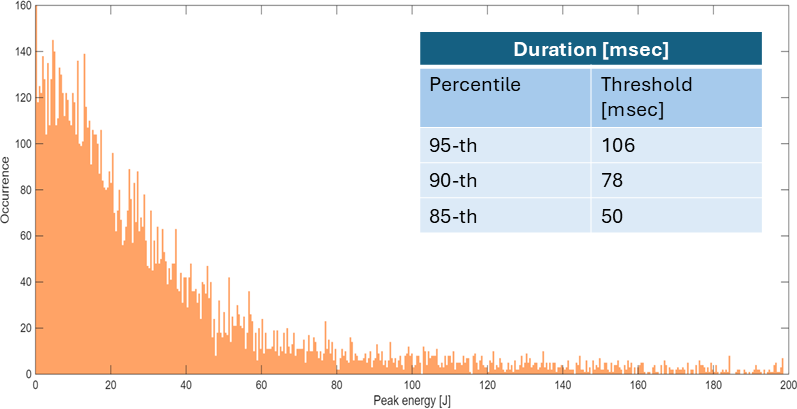}
\caption{Characteristics of power spikes identified in the real life datasets of AI power trends: histogram of peak energy (expressed in Joule), and summary of the thresholds of the percentiles of the peak durations (top right).}
\label{fig_Stats}
\end{figure}

This analysis helps us to appreciate that the vast majority (i.e., between 85 and 95$\%$) of the power spikes lasts at most 100 msec (figure~\ref{fig_Stats}). 
It is important to note that this result is biased by the sensing capacity of the datasets that have been considered (which spans between 3 and 100 msec). 
Therefore, it is possible to assume that shorter power spikes could occur as well. 

The very short duration of the power spikes comes with low energy load. % that each spike provides. 
Indeed, across the various values of thresholds that we considered, the energy consumed by each spike we identified falls within 5 and 100 J, with a mean centered approximately at 50 J (see figure~\ref{fig_Stats}). 
At the same time, the instantaneous power consumption of each peak spans from 1$\%$ to 5$\%$ of the maximum power of each rack for over 85$\%$ of the spikes. 

At this point, it is important not to underestimate the impact of the power spikes onto the working conditions of AI data centres. 
In fact, although the power spikes in the time series of power draw could seem limited and modest, the aforementioned results are calculated per rack. 
This is not a negligible detail: indeed, the number of racks in a classic AI data centre typically sits between 1000 and 1200. 
Thus, the actual impact of these power fluctuations on the data centre systems could be orders of magnitude higher at each moment~\cite{MSOpenAINvidia,PowerAI,Alberta,IEA,Verne_AIheat,JLL}. 
These figures hence illustrate to the extent of the power capacity oversizing should be implemented to ensure the smooth functioning of the AI data centres when considering the use of low dynamic power resources as input to the computing systems. 

Also, sudden power fluctuations can influence the robustness and effectiveness of AI data centre service. 
Therefore, it is crucial to identify solutions that can make the $P_{ext}$ term in (\ref{eq_power_balance}) so highly dynamic that it absorbs all of the spikes %to absorb all the spikes
above the threshold imposed by the power grid working conditions -  this operation is called ``power shaving''. 

\begin{figure}[h]
\centering
\includegraphics[width=1\textwidth]{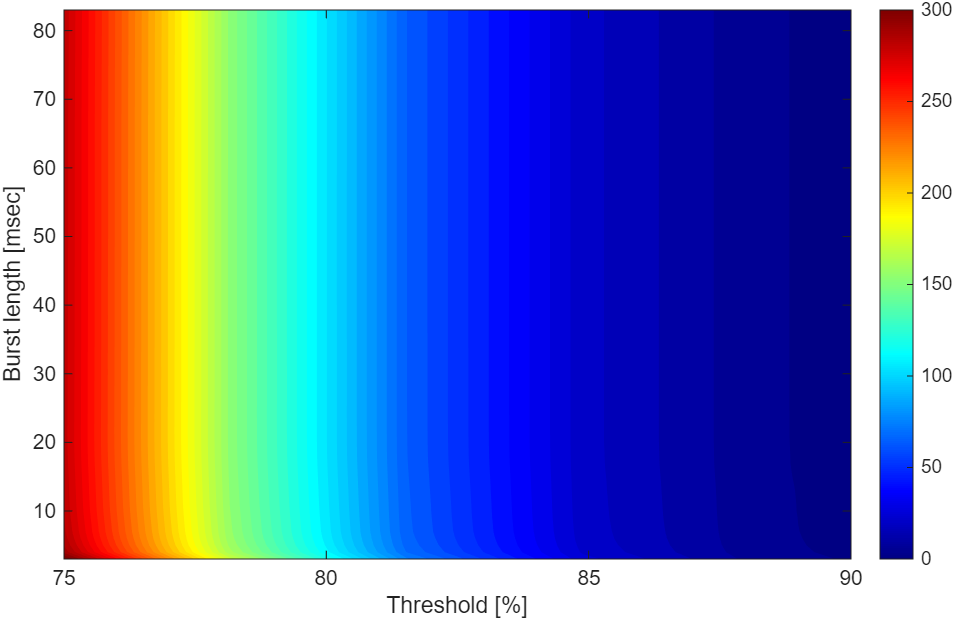}
\caption{Number of GPUs that would not face shutdown when power spikes occurring of length greater than 'Burst length' over the limit of 'Threshold' percentage of the rack maximum power could be absorbed. 
We assumed each GPU to be modeled around the Nvidia H100 model, i.e., showing instantaneous power draw of 700W when on training phase.}
\label{fig_GPUsave}
\end{figure}

To assess the effect of the power spikes, we quantified the number of AI accelerators that could be saved from shutdown or interruption per rack assuming that sudden fluctuations induced by power spikes could be addressed and absorbed by additional power systems associated to the AI data centres (i.e., that could be modeled with an highly dynamic $P_{ext}$ term in (\ref{eq_power_balance}). 
These results are displayed in Figure~\ref{fig_GPUsave}, where we report the number of GPUs that would not face shutdown when power spikes occurring of length greater than 'Burst length' over the limit of 'Threshold' percentage of the rack maximum power could be absorbed. 
For this computation we assumed each GPU to be modeled around the Nvidia H100 model, i.e., showing instantaneous power draw of 700W when on training phase. 
Also, it is important to consider that these results are obtained in an  ideal situation for power shaving (i.e., all the power spikes can be absorbed), hence the outcomes in Figure~\ref{fig_GPUsave} represent an upper limit for the dynamic power response performance. 

Taking a look to the extent of the results in Figure~\ref{fig_GPUsave}, 
it is possible to appreciate the %several comments can be drawn. 
%On one side, these outcomes show 
the massive impact of sudden power fluctuations onto the AI data centre working conditions. 
In fact, to provide a metric by rule of thumb for this, we can consider that 200 GPUs can be located in one rack in an AI hyperscaler centre. 
Therefore, we can state that discarding  the effect of power spikes could lead in principle to catastrophic consequences for the AI data centres.

%This estimate 
%has at least two direct consequences. 
%On one hand,  since energy contribution is pretty modest, it is possible 
%to consider 
In order to achieve the results displayed in Figure~\ref{fig_GPUsave}, several hardware solutions could be put in place~\cite{MSOpenAINvidia}, which can be categorised in three main classes with respect to their technical implementation: 
\begin{itemize}
    \item  The limited energy contribution of the spikes enables the implementation of \underline{\textbf{passive devices}} (e.g., capacitors) connected to the power input per AI accelerator and rack to smoothen out these sudden power surges. These devices would be able to provide a very limited amount of charge (stored while the AI accelerator are not working at peak usage), that can be used once a peak would be detected in the power draw. These devices would then recharge as soon as the power draw would fall under a specific threshold of the power consumption. Therefore, these devices would act as external sources of energy (albeit minimal), thus enabling the $P_{ext}$ in (\ref{eq_power_balance}) to become dynamic. 
    
    \item The highly variable and unpredictable pattern of power spikes in AI accelerators would demand connecting independent power sources (e.g., battery energy storage systems) to the power grid.     
    These devices become effective instruments for efficient peak shaving and power response by actively switching modes between charging and discharging (i.e., making them \underline{\textbf{active devices}}). 
    The charge and discharge phases (occurring during low power communication and high-power computation intervals, respectively) of this type of devices can be directly translated in terms of the dynamics of the $P_{ext}$ in (\ref{eq_power_balance}).

\end{itemize}

When comparing the performance of passive and active devices for dynamic power response, it is important to assess the 
%Although very simple in design and execution, the use of passive devices to enable dynamic power response might not fulfill 
conditions for optimising the supply and demand conditions required by AI data centres. 
Specifically, these conditions can be summarised by considering:
\begin{itemize}
    \item computational gain;
    \item ability to avoid the use of artificial loads when AI accelerators are not in use for computing (``dummy loads'');
    \item CAPEX reduction;
    \item management costs.
\end{itemize}

Table~\ref{tab_comparison} reports a comparison across the main strategies to implement power dynamic response for AI data centres. The best results for an efficient and sustainable functioning of AI data centres are highlighted in blue, whilst the critical factors are written in red. 

\begin{table}[h]
\caption{Performance comparison between solutions for dynamic power response}\label{tab_comparison}%
\begin{tabular}{@{}lllll@{}}
\toprule
\textbf{Strategy} & \textbf{Computational gain}  & \textbf{Dummy loads} & \textbf{CAPEX reduction} & \textbf{Management costs}\\
\midrule
Passive (capacitors)   & \color{blue}+100 $\%$ \color{black}   & \color{red} Yes \color{black} & \color{red} - \color{black}  & \color{red} High \color{black}\\
Passive (supercapacitors)   & \color{red}+40 $\%$ \color{black}  & \color{blue} No \color{black}  & \color{red} +45$\%$ \color{black} & \color{red} High \color{black}\\
Active   & \color{blue}+100 $\%$ \color{black}   & \color{blue} No \color{black}  & \color{blue} +55$\%$ \color{black} & \color{blue} Low \color{black} \\
\botrule
\end{tabular}
%\footnotetext{Source: This is an example of table footnote. This is an example of table footnote.}
%\footnotetext[1]{Example for a first table footnote. This is an example of table footnote.}
%\footnotetext[2]{Example for a second table footnote. This is an example of table footnote.}
\end{table}

As previously mentioned, avoiding service disruptions and AI accelerators shutdown is one of the main concerns to be addressed to guarantee efficiency of the data centres. 
In this respect, peak shaving systems, either directed by means of passive or active strategies for dynamic power allocation, could make the difference in the everyday operations of the data centres. 
However, the approach to achieve this result might have an effect on other aspects of the AI data centres operations.

For instance, passive solutions for dynamic power response would typically show slow responsiveness to the deeper transitions on AI accelerator usage, e.g., moving from peak usage to idle (and viceversa). 
%
%
%imply capacitative devices that show slow dynamic range, especially when considering the transitions from full AI accelerator usage to idle phases. 
This means that power would be consumed also during intervals when the AI accelerator should not be working for AI model computation, especially to maange the transitions between peak usage and idle (``ramp down''), and viceversa (``ramp up'')  (see Figure~\ref{fig_Dummy}(B,C)), hence cutting the efficiency of AI data centres by more than 50$\%$~\cite{Nvidia_capacitor,MSOpenAINvidia,IEA,McKinsey,Skeleton_GPUGraphene}. 
Also, this would be detrimental to the reliability and robustness of these solutions, making them struggle to ensure the performance demands in high computational stress conditions~\cite{MSOpenAINvidia}. 

On one hand, this implies higher energy consumption, directly translating into higher financial commitments to guarantee the power supply of the data centres (to be estimated in millions of US dollars per year per data centre~\cite{McKinsey,McKinsey2,JLL,Skeleton_GPUGraphene,IEA,Verne_AIheat}). 
On the other hand, dummy loads means that temperature profiles within the racks and the shelves would not have the possibility to relax on the long term. 
This hinders the computational capacity of the AI data centres, as thermal overflow affects the integrated circuitry of the AI accelerators by reducing their life time, hence the reliability of the AI models~\cite{TAPAS_Microsoft,Microsoft24,Alberta,MSOpenAINvidia}. 

It is true that passive solutions could overcome these issues by considering more sophisticated devices (e.g., supercapacitors~\cite{Skeleton_GPUGraphene}), that would make the AI accelerator power trends move from Figure~\ref{fig_Dummy}(B) to the shape of Figure~\ref{fig_Dummy}(C). 
However, this typically comes as a cost on the computational gain, since the performance of peak shaving process might be suboptimal. 

Active solutions would instead avoid all the aforesaid problems, hence leading to the maximisation of the computational gain and minimisation of the energy consumption. 
In turn, this means obtaining a stronger CAPEX reduction, thus improving the efficiency of the capital investments in the management of AI data centres. 
To achieve these goals, active solutions must be able to track extreme fluctuations (i.e., variations at high frequency) while supporting large capacitance for energy storage~\cite{MSOpenAINvidia}. In this respect, ultra-fast charging systems (e.g., based on Li-ion cells \cite{ref_nyobolt}) could be key solutions to achieve the efficiency, robustness, and reliability goals demanded to make AI data centres sustainable and financially viable~\cite{MSOpenAINvidia}. 

It is also worth noting that active solutions could facilitate the implementation of management strategies to improve the efficiency of AI data centres. 
In fact, active solutions for dynamic power response can be implemented per rack (i.e., approximately 200 AI accelerators), whilst passive solutions would require devices to be connected to each AI accelerator. 
This means that active solutions would enable easier algorithmic strategies for coordinating the dynamic power response system with the other components of the AI data centres. 
For instance, scheduling methods for intelligent distribution of computational load, power and temperature (e.g.,~\cite{TAPAS_Microsoft}) would help in further improving the system efficiency, reducing the randomness produced by the irregular access to the data centres, as well as the type of jobs required by the users. 

%\color{blue}  \textbf{GREEN AI } \color{black}

The impact of the development and implementation of dynamic power response systems would span across several diverse sectors of society, affecting the sustainability of AI in the next decades~\cite{SustainableAI,Science_masanet,NatClimChange_Masanet,NatClimChange_Rolnick,NPJ_water}. 
In particular, we can mention the following:
\begin{itemize}
    \item improving the efficiency of data centres have direct effects on the computational, financial and structural planning of AI hyperscalers. 
    In fact, it allows providing more than 50$\%$ of the actual computational power delivered by AI accelerators, without imposing additional operational costs for the AI hyperscalers managers~\cite{Alberta,McKinsey,MSOpenAINvidia,Skeleton_GPUGraphene}.  This entails also the costs for infrastructures, hence reducing the use of new land for the construction of new data centres~\cite{McKinsey2}. This translates into reducing the need for demand of gas- and coal-fired power plants to support the AI data centre growth~\cite{IEA,Bloomberg}, which would be detrimental to meeting the objectives of sustainable development and greenhouse gas emissions aiming to mitigate climate change effects~\cite{IEA,UNFCCC}; 
    \item enhancing the power consumption of AI data centres enables the development of robust algorithms to actually implement the transition to \textit{green AI}~\cite{Alberta,MSOpenAINvidia,TAPAS_Microsoft}. In fact, ensuring power continuity (without service disruptions) allows the deployment of AI strategies that guarantee functionality also in case of unstable power conditions and fluctuations of the AI queries. Hence, data centres would not become bottlenecks to the development of AI architectures demanding for less computational power, thus supporting the increase of sustainability of AI models in modern society.
    \item efficient AI data centres result in less heat release in the atmosphere, therefore reducing their impact on environment and sustainable development, especially considering the  scenarios entailed by the shared socioeconomic pathways (SSPs) for the next decades~\cite{McKinsey,McKinsey2,JLL,Verne_AIheat}. Decreasing the heat waste generated by AI data centres have especially a direct impact on local communities and the local climate zones of the areas surrounding their infrastructures~\cite{IEA}. Also, making AI data centres more efficient and robust to power fluctuations leads to reduction of carbon emissions and water usage~\cite{McKinsey2,IEA, Bloomberg}. 
    Ultimately, solutions for dynamic power response (especially active ones) could turn AI data centers from causes of distortions in the power grid to stabilisers, eventually removing AI hyperscalers from the  grid during periods of high stress (e.g., hot summer nights) ~\cite{Bloomberg,MSOpenAINvidia,Alberta,McKinsey2}; 
    %serve as stabilizers of power grid of the regions around the AI hyperscalers
    %\textbf{LESS HEAT, MORE STABILITY FOR GRID --> help for communities} 
    \item fostering the design of hardware-aware AI models that dynamically adjust computation, memory usage, and precision based on real-time power availability, thus synergizing algorithmic efficiency with infrastructure-level energy management to further enhance the sustainability and resilience of AI systems. Indeed,  beyond passive and active hardware solutions, an emerging paradigm involves the use of intelligent control systems that integrate hardware-aware AI models with real-time power management algorithms~\cite{pantow,pansel}. These systems dynamically coordinate computation scheduling, model precision, and accelerator usage in response to instantaneous power availability and thermal conditions. By coupling algorithmic adaptivity with infrastructure-level monitoring, intelligent control systems effectively close the loop between AI workloads and power delivery, optimizing both performance and energy efficiency in dynamic operating environments.
\end{itemize}

%\color{blue}
%\textbf{EXPLANATION ON IMPACT (TEMPERATURE, FINANCIAL, scheduling)}
%\color{black}

\section{Conclusion}

The growth of AI data centres implies a huge impact on power infrastructures and sustainable development. 
Making part of the data centre input power resources highly dynamic would induce a number of advantages, i.e., to reduce data centre downtime, protect infrastructure by power fluctuations, drops and spikes, and enable the data centre structures more resilient towards irregular and unstructured AI platform usage. 
The main benefits of this paradigm shift can be categorized as follows:
\begin{itemize}
    \item very rapid power delivery to absorb sharp, short, and high energy spikes;
    \item reduction of energy consumption and reduced operating costs, flattening the energy demand curve and eliminating the need for artificial, empty compute ``dummy loads''; %, which passive solutions (e.g., capacitors) cannot avoid;
    \item increasing reliability, and reducing the stress on equipment and failures of AI accelerators; 
    \item reducing the need to oversize backup generators, batteries, or transformers, leading to smarter CAPEX allocation and lower maintenance; 
    \item developing hardware-aware AI models that adapt computational load and precision dynamically to available power and thermal budgets, improving energy efficiency and sustainability.
\end{itemize}

Active and passive solutions can be implemented to achieve these outcomes. 
In this work, the capacity and limits of the main techiniques for dynamic power response have been investigated and compared, and the direct effects and their implications have been discussed on demand, supply and sustainable development. 
These benefits could be maximised by implementing additional algorithmic strategies for power-load-temperature scheduling, efficient cooling infrastructures (e.g., air or liquid), and alternative input power sources (e.g., use of renewable energy).

%highly dynamic power profiles, causing short, unpredictable spikes in demand

%distrurbance of the service. 
%To avoid catastrophic disruption of the data analysis systems, AI companies are focusing big parts of their budget towards the development of robust AI infrastructures and reliable power services. 
%This is because the system inefficiency induced by the stress imposed by AI-based data analysis on power infrastructures might have repercussions across the demand, supply 

%\color{red} \textbf{CUT HERE} 
%\color{black}

\bibliography{sn-bibliography}


\begin{thebibliography}{35}
\ifx \bisbn   \undefined \def \bisbn  #1{ISBN #1}\fi
\ifx \binits  \undefined \def \binits#1{#1}\fi
\ifx \bauthor  \undefined \def \bauthor#1{#1}\fi
\ifx \batitle  \undefined \def \batitle#1{#1}\fi
\ifx \bjtitle  \undefined \def \bjtitle#1{#1}\fi
\ifx \bvolume  \undefined \def \bvolume#1{\textbf{#1}}\fi
\ifx \byear  \undefined \def \byear#1{#1}\fi
\ifx \bissue  \undefined \def \bissue#1{#1}\fi
\ifx \bfpage  \undefined \def \bfpage#1{#1}\fi
\ifx \blpage  \undefined \def \blpage #1{#1}\fi
\ifx \burl  \undefined \def \burl#1{\textsf{#1}}\fi
\ifx \doiurl  \undefined \def \doiurl#1{\url{https://doi.org/#1}}\fi
\ifx \betal  \undefined \def \betal{\textit{et al.}}\fi
\ifx \binstitute  \undefined \def \binstitute#1{#1}\fi
\ifx \binstitutionaled  \undefined \def \binstitutionaled#1{#1}\fi
\ifx \bctitle  \undefined \def \bctitle#1{#1}\fi
\ifx \beditor  \undefined \def \beditor#1{#1}\fi
\ifx \bpublisher  \undefined \def \bpublisher#1{#1}\fi
\ifx \bbtitle  \undefined \def \bbtitle#1{#1}\fi
\ifx \bedition  \undefined \def \bedition#1{#1}\fi
\ifx \bseriesno  \undefined \def \bseriesno#1{#1}\fi
\ifx \blocation  \undefined \def \blocation#1{#1}\fi
\ifx \bsertitle  \undefined \def \bsertitle#1{#1}\fi
\ifx \bsnm \undefined \def \bsnm#1{#1}\fi
\ifx \bsuffix \undefined \def \bsuffix#1{#1}\fi
\ifx \bparticle \undefined \def \bparticle#1{#1}\fi
\ifx \barticle \undefined \def \barticle#1{#1}\fi
\bibcommenthead
\ifx \bconfdate \undefined \def \bconfdate #1{#1}\fi
\ifx \botherref \undefined \def \botherref #1{#1}\fi
\ifx \url \undefined \def \url#1{\textsf{#1}}\fi
\ifx \bchapter \undefined \def \bchapter#1{#1}\fi
\ifx \bbook \undefined \def \bbook#1{#1}\fi
\ifx \bcomment \undefined \def \bcomment#1{#1}\fi
\ifx \oauthor \undefined \def \oauthor#1{#1}\fi
\ifx \citeauthoryear \undefined \def \citeauthoryear#1{#1}\fi
\ifx \endbibitem  \undefined \def \endbibitem {}\fi
\ifx \bconflocation  \undefined \def \bconflocation#1{#1}\fi
\ifx \arxivurl  \undefined \def \arxivurl#1{\textsf{#1}}\fi
\csname PreBibitemsHook\endcsname

%%% 1
\bibitem[\protect\citeauthoryear{Li et~al.}{2024}]{Alberta}
\begin{botherref}
\oauthor{\bsnm{Li}, \binits{Y.}},
\oauthor{\bsnm{Mughees}, \binits{M.}},
\oauthor{\bsnm{Chen}, \binits{Y.}},
\oauthor{\bsnm{Li}, \binits{Y.R.}}:
The Unseen AI Disruptions for Power Grids: LLM-Induced Transients.
Preprint at \url{https://arxiv.org/abs/2409.11416}
(2024)
\end{botherref}
\endbibitem

%%% 2
\bibitem[\protect\citeauthoryear{Chen et~al.}{2025}]{PowerAI}
\begin{barticle}
\bauthor{\bsnm{Chen}, \binits{M.}}, \betal:
\batitle{Power for ai and ai for power: The infinite entanglement between artificial intelligence and power electronics systems}.
\bjtitle{IEEE Power Electronics Magazine}
\bvolume{12}(\bissue{1}),
\bfpage{37}--\blpage{43}
(\byear{2025})
\end{barticle}
\endbibitem

%%% 3
\bibitem[\protect\citeauthoryear{Mytton}{2021}]{NPJ_water}
\begin{botherref}
\oauthor{\bsnm{Mytton}, \binits{D.}}:
Data centre water consumption.
npj Clean Water
\textbf{4}(1)
(2021)
\end{botherref}
\endbibitem

%%% 4
\bibitem[\protect\citeauthoryear{Masanet et~al.}{2013}]{NatClimChange_Masanet}
\begin{barticle}
\bauthor{\bsnm{Masanet}, \binits{E.}},
\bauthor{\bsnm{Shehabi}, \binits{A.}},
\bauthor{\bsnm{Koomey}, \binits{J.G.}}:
\batitle{Characteristics of low-carbon data centres}.
\bjtitle{Nature Clim Change}
\bvolume{3}(\bissue{7}),
\bfpage{627}--\blpage{630}
(\byear{2013})
\end{barticle}
\endbibitem

%%% 5
\bibitem[\protect\citeauthoryear{Kaack et~al.}{2022}]{NatClimChange_Rolnick}
\begin{botherref}
\oauthor{\bsnm{Kaack}, \binits{L.H.}}, et al.:
Aligning artificial intelligence with climate change mitigation.
Nature Clim Change
\textbf{12}(6)
(2022)
\end{botherref}
\endbibitem

%%% 6
\bibitem[\protect\citeauthoryear{Masanet et~al.}{2020}]{Science_masanet}
\begin{barticle}
\bauthor{\bsnm{Masanet}, \binits{E.}}, \betal:
\batitle{Recalibrating global data center energy-use estimates}.
\bjtitle{Science}
\bvolume{367}(\bissue{6481}),
\bfpage{984}--\blpage{986}
(\byear{2020})
\end{barticle}
\endbibitem

%%% 7
\bibitem[\protect\citeauthoryear{Verne}{2024}]{Verne_AIheat}
\begin{botherref}
\oauthor{\bsnm{Verne}}:
Harnessing Data Center Waste Heat.
Available at \url{https://www.verneglobal.com/blog/data-center-waste-heat}
(2024)
\end{botherref}
\endbibitem

%%% 8
\bibitem[\protect\citeauthoryear{Steele}{2025}]{JLL}
\begin{botherref}
\oauthor{\bsnm{Steele}, \binits{K.}}:
Global data center demand surges despite supply and power constraints.
Available at \url{https://www.jll.com/en-us/newsroom/global-data-center-demand-surges-despite-supply-and-power-constraints}
(2025)
\end{botherref}
\endbibitem

%%% 9
\bibitem[\protect\citeauthoryear{McKinsey}{2024}]{McKinsey}
\begin{botherref}
\oauthor{\bsnm{McKinsey}}:
How data centers and the energy sector can sate AI's hunger for power.
Available at \url{https://www.mckinsey.com/industries/private-capital/our-insights/how-data-centers-and-the-energy-sector-can-sate-ais-hunger-for-power}
(2024)
\end{botherref}
\endbibitem

%%% 10
\bibitem[\protect\citeauthoryear{McKinsey}{2025}]{McKinsey2}
\begin{botherref}
\oauthor{\bsnm{McKinsey}}:
Scaling bigger, faster, cheaper data centers with smarter designs.
Available at \url{https://www.mckinsey.com/industries/private-capital/our-insights/scaling-bigger-faster-cheaper-data-centers-with-smarter-designs}
(2025)
\end{botherref}
\endbibitem

%%% 11
\bibitem[\protect\citeauthoryear{Desislavov et~al.}{2023}]{AIEnergyConsum}
\begin{barticle}
\bauthor{\bsnm{Desislavov}, \binits{R.}},
\bauthor{\bsnm{Mart{\'\i}nez-Plumed}, \binits{F.}},
\bauthor{\bsnm{Hern{\'a}ndez-Orallo}, \binits{J.}}:
\batitle{Trends in ai inference energy consumption: Beyond the performance-vs-parameter laws of deep learning}.
\bjtitle{Sustainable Computing: Informatics and Systems}
\bvolume{38},
\bfpage{100857}
(\byear{2023})
\end{barticle}
\endbibitem

%%% 12
\bibitem[\protect\citeauthoryear{Siddik et~al.}{2021}]{DataCentre_footprint}
\begin{botherref}
\oauthor{\bsnm{Siddik}, \binits{M.A.B.}},
\oauthor{\bsnm{Shehabi}, \binits{A.}},
\oauthor{\bsnm{Marston}, \binits{L.}}:
The environmental footprint of data centers in the united states.
Environ. Res. Lett.
\textbf{16}(6)
(2021)
\end{botherref}
\endbibitem

%%% 13
\bibitem[\protect\citeauthoryear{Bouza et~al.}{2023}]{DataCentre_footprint2}
\begin{botherref}
\oauthor{\bsnm{Bouza}, \binits{L.}},
\oauthor{\bsnm{Bugeau}, \binits{A.}},
\oauthor{\bsnm{Lannelongue}, \binits{L.}}:
How to estimate carbon footprint when training deep learning models? a guide and review.
Environ. Res. Commun.
\textbf{5}(11)
(2023)
\end{botherref}
\endbibitem

%%% 14
\bibitem[\protect\citeauthoryear{Patel et~al.}{2024}]{DataCentreDilemma}
\begin{botherref}
\oauthor{\bsnm{Patel}, \binits{D.}},
\oauthor{\bsnm{Nishball}, \binits{D.}},
\oauthor{\bsnm{Ontiveros}, \binits{J.E.}}:
AI Datacenter Energy Dilemma - Race for AI Datacenter Space.
Available at \url{https://www.semianalysis.com/p/ai-datacenter-energy-dilemma-race}
(2024)
\end{botherref}
\endbibitem

%%% 15
\bibitem[\protect\citeauthoryear{Luccioni et~al.}{2024}]{PowerHungry}
\begin{bchapter}
\bauthor{\bsnm{Luccioni}, \binits{A.S.}},
\bauthor{\bsnm{Jernite}, \binits{Y.}},
\bauthor{\bsnm{Strubell}, \binits{E.}}:
\bctitle{Power hungry processing: Watts driving the cost of ai deployment?}
In: \beditor{\bsnm{ACM}} (ed.)
\bbtitle{Proceedings of Conference on Fairness, Accountability, and Transparency}
(\byear{2024})
\end{bchapter}
\endbibitem

%%% 16
\bibitem[\protect\citeauthoryear{Choukse et~al.}{2025}]{MSOpenAINvidia}
\begin{botherref}
\oauthor{\bsnm{Choukse}, \binits{E.}}, et al.:
Power Stabilization for AI Training Datacenters.
Preprint at \url{https://arxiv.org/abs/2508.14318}
(2025)
\end{botherref}
\endbibitem

%%% 17
\bibitem[\protect\citeauthoryear{Amodei et~al.}{2025}]{OpenAI_compute}
\begin{botherref}
\oauthor{\bsnm{Amodei}, \binits{D.}}, et al.:
AI and compute.
Available at \url{https://openai.com/index/ai-and-compute/}
(2025)
\end{botherref}
\endbibitem

%%% 18
\bibitem[\protect\citeauthoryear{Wu et~al.}{2022}]{SustainableAI}
\begin{bchapter}
\bauthor{\bsnm{Wu}, \binits{C.-J.}}, \betal:
\bctitle{Sustainable ai: Environmental implications, challenges and opportunities}.
In: \bbtitle{Proceedings of Machine Learning and Systems}
(\byear{2022})
\end{bchapter}
\endbibitem

%%% 19
\bibitem[\protect\citeauthoryear{Govind et~al.}{2023}]{SuperComputing}
\begin{bchapter}
\bauthor{\bsnm{Govind}, \binits{A.}}, \betal:
\bctitle{Comparing power signatures of hpc workloads: Machine learning vs simulation}.
In: \beditor{\bsnm{ACM}} (ed.)
\bbtitle{Proceedings of SC '23 Workshops of The International Conference on High Performance Computing, Network, Storage, and Analysis}
(\byear{2023})
\end{bchapter}
\endbibitem

%%% 20
\bibitem[\protect\citeauthoryear{Stojkovic et~al.}{2025}]{TAPAS_Microsoft}
\begin{bchapter}
\bauthor{\bsnm{Stojkovic}, \binits{J.}}, \betal:
\bctitle{Tapas: Thermal- and power-aware scheduling for llm inference in cloud platforms}.
In: \beditor{\bsnm{ACM}} (ed.)
\bbtitle{Proceedings of 30th ACM International Conference on Architectural Support for Programming Languages and Operating Systems}
(\byear{2025})
\end{bchapter}
\endbibitem

%%% 21
\bibitem[\protect\citeauthoryear{Patel et~al.}{2024}]{Microsoft24}
\begin{bchapter}
\bauthor{\bsnm{Patel}, \binits{P.}}, \betal:
\bctitle{Characterizing power management opportunities for llms in the cloud}.
In: \beditor{\bsnm{ACM}} (ed.)
\bbtitle{Proceedings of 29th ACM International Conference on Architectural Support for Programming Languages and Operating Systems}
(\byear{2024})
\end{bchapter}
\endbibitem

%%% 22
\bibitem[\protect\citeauthoryear{Delestrac et~al.}{2024}]{GPUEnergyConsum}
\begin{bchapter}
\bauthor{\bsnm{Delestrac}, \binits{P.}}, \betal:
\bctitle{Analyzing gpu energy consumption in data movement and storage}.
In: \beditor{\bsnm{IEEE}} (ed.)
\bbtitle{Proceedings of International Conference on Application-specific Systems, Architectures and Processors (ASAP)}
(\byear{2024})
\end{bchapter}
\endbibitem

%%% 23
\bibitem[\protect\citeauthoryear{Newkirk et~al.}{2025}]{Brookhaven}
\begin{botherref}
\oauthor{\bsnm{Newkirk}, \binits{A.}}, et al.:
Empirically-Calibrated H100 Node Power Models for Reducing Uncertainty in AI Training Energy Estimation.
Preprint at \url{https://arxiv.org/pdf/2506.14551v1}
(2025)
\end{botherref}
\endbibitem

%%% 24
\bibitem[\protect\citeauthoryear{Narayanan et~al.}{2021}]{LLM_megatron}
\begin{bchapter}
\bauthor{\bsnm{Narayanan}, \binits{D.}}, \betal:
\bctitle{Efficient large-scale language model training on gpu clusters using megatron-lm}.
In: \beditor{\bsnm{ACM}} (ed.)
\bbtitle{Proceedings of the International Conference for High Performance Computing, Networking, Storage and Analysis}
(\byear{2021})
\end{bchapter}
\endbibitem

%%% 25
\bibitem[\protect\citeauthoryear{IEA}{2025}]{IEA}
\begin{botherref}
\oauthor{\bsnm{IEA}}:
Energy and AI.
Available at \url{https://www.iea.org/reports/energy-and-ai}
(2025)
\end{botherref}
\endbibitem

%%% 26
\bibitem[\protect\citeauthoryear{Nicoletti et~al.}{2024}]{Bloomberg}
\begin{botherref}
\oauthor{\bsnm{Nicoletti}, \binits{L.}},
\oauthor{\bsnm{Malik}, \binits{N.}},
\oauthor{\bsnm{Tartar}, \binits{A.}}:
AI needs so much power, it's making yours worse.
Available at \url{https://www.bloomberg.com/graphics/2024-ai-power-home-appliances/}
(2024)
\end{botherref}
\endbibitem

%%% 27
\bibitem[\protect\citeauthoryear{Schneider et~al.}{2025}]{LCA_AI}
\begin{botherref}
\oauthor{\bsnm{Schneider}, \binits{I.}}, et al.:
Life-Cycle Emissions of AI Hardware: A Cradle-To-Grave Approach and Generational Trends.
Preprint at \url{https://arxiv.org/abs/2502.01671}
(2025)
\end{botherref}
\endbibitem

%%% 28
\bibitem[\protect\citeauthoryear{Dimitrov et~al.}{2025}]{Nvidia_capacitor}
\begin{botherref}
\oauthor{\bsnm{Dimitrov}, \binits{R.}}, et al.:
How New GB300 NVL72 Features Provide Steady Power for AI.
Available at \url{https://developer.nvidia.com/blog/how-new-gb300-nvl72-features-provide-steady-power-for-ai/}
(2025)
\end{botherref}
\endbibitem

%%% 29
\bibitem[\protect\citeauthoryear{Skeleton}{2025}]{Skeleton_GPUGraphene}
\begin{botherref}
\oauthor{\bsnm{Skeleton}}:
Maximize the full power of GPUs in AI data centers.
Available at \url{https://www.skeletontech.com/maximize-gpu-power-in-ai-data-centers}
(2025)
\end{botherref}
\endbibitem

%%% 30
\bibitem[\protect\citeauthoryear{Samsi et~al.}{2021}]{MIT_supercloud}
\begin{bchapter}
\bauthor{\bsnm{Samsi}, \binits{S.}}, \betal:
\bctitle{The mit supercloud dataset}.
In: \beditor{\bsnm{IEEE}} (ed.)
\bbtitle{Proceedings of High Performance Extreme Computing Conference (HPEC)}
(\byear{2021})
\end{bchapter}
\endbibitem

%%% 31
\bibitem[\protect\citeauthoryear{Sakalkar et~al.}{2020}]{Google_dataset}
\begin{bchapter}
\bauthor{\bsnm{Sakalkar}, \binits{V.}}, \betal:
\bctitle{Data center power oversubscription with a medium voltage power plane and priority-aware capping}.
In: \beditor{\bsnm{ACM}} (ed.)
\bbtitle{Proceedings of International Conference on Architectural Support for Programming Languages and Operating Systems, Association for Computing Machinery}
(\byear{2020})
\end{bchapter}
\endbibitem

%%% 32
\bibitem[\protect\citeauthoryear{Shivareddy}{2025}]{ref_nyobolt}
\begin{botherref}
\oauthor{\bsnm{Shivareddy}, \binits{S.}}:
Maximising AI performance in data centers with dynamic power response systems.
Available at \url{https://https://nyobolt.com/resources/blog/maximising-ai-performance-in-data-centers-with-dynamic-power-response-systems/}
(2025)
\end{botherref}
\endbibitem

%%% 33
\bibitem[\protect\citeauthoryear{UNFCCC}{2024}]{UNFCCC}
\begin{botherref}
\oauthor{\bsnm{UNFCCC}}:
United Nations Framework Convention on Climate Change 2024 nationally determined contributions (NDC) Synthesis Report.
Available at \url{https://unfccc.int/process-and-meetings/the-paris-agreement/nationally-determined-contributions-ndcs/2024-ndc-synthesis-report}
(2024)
\end{botherref}
\endbibitem

%%% 34
\bibitem[\protect\citeauthoryear{Pandelea et~al.}{2022}]{pantow}
\begin{barticle}
\bauthor{\bsnm{Pandelea}, \binits{V.}},
\bauthor{\bsnm{Ragusa}, \binits{E.}},
\bauthor{\bsnm{Young}, \binits{T.}},
\bauthor{\bsnm{Gastaldo}, \binits{P.}},
\bauthor{\bsnm{Cambria}, \binits{E.}}:
\batitle{Toward hardware-aware deep-learning-based dialogue systems}.
\bjtitle{Neural Computing and Applications}
\bvolume{34},
\bfpage{10397}--\blpage{10408}
(\byear{2022})
\end{barticle}
\endbibitem

%%% 35
\bibitem[\protect\citeauthoryear{Pandelea et~al.}{2023}]{pansel}
\begin{bchapter}
\bauthor{\bsnm{Pandelea}, \binits{V.}},
\bauthor{\bsnm{Ragusa}, \binits{E.}},
\bauthor{\bsnm{Gastaldo}, \binits{P.}},
\bauthor{\bsnm{Cambria}, \binits{E.}}:
\bctitle{Selecting Language Models Features Via Software-Hardware Co-Design}.
In: \bbtitle{Proceedings of {IEEE ICASSP}}
(\byear{2023})
\end{bchapter}
\endbibitem

\end{thebibliography}


\end{document}